\documentclass[11pt,a4paper]{article}
\usepackage[hyperref]{acl2019}
\usepackage{times}
\usepackage{latexsym}
\usepackage{paralist}
\usepackage{url}
\usepackage{multirow}
\usepackage{enumerate}
\usepackage[shortlabels]{enumitem}

\aclfinalcopy 

\title{Deep Neural Models for Medical Concept Normalization in User-Generated Texts}

\author{Zulfat Miftahutdinov \\
  Kazan Federal University,\\
  Kazan, Russia \\
  \texttt{zulfatmi@gmail.com} \\\And
  Elena Tutubalina \\
  Kazan Federal University,\\
  Kazan, Russia \\
  Samsung-PDMI Joint AI Center,\\ 
  PDMI RAS, St. Petersburg, Russia \\
  \texttt{elvtutubalina@kpfu.ru} 
  \\}

\date{}

\begin{document}
\maketitle
\begin{abstract}
In this work, we consider the \emph{medical concept normalization} problem, i.e., the problem of mapping a health-related entity mention in a free-form text to a concept in a controlled vocabulary, usually to the standard thesaurus in the Unified Medical Language System (UMLS).
This is a challenging task since medical terminology is very different when coming from health care professionals or from the general public in the form of social media texts. We approach it as a sequence learning problem with powerful neural networks such as recurrent neural networks and contextualized word representation models trained to obtain semantic representations of social media expressions. Our experimental evaluation over three different benchmarks shows that neural architectures leverage the semantic meaning of the entity mention and significantly outperform an existing state of the art models.
\end{abstract}

\section{Introduction}
User-generated texts (UGT) on social media present a wide variety of facts, experiences, and opinions on numerous topics, and this treasure trove of information is currently severely underexplored. We consider the problem of discovering medical concepts in UGTs with the ultimate goal of mining new symptoms, adverse drug reactions (ADR), and other information about a disorder or a drug. 

An important part of this problem is to translate a text from ``social media language'' (e.g., ``can't fall asleep all night'' or ``head spinning a little'') to ``formal medical language'' (e.g., ``insomnia'' and ``dizziness'' respectively). This is necessary to match user-generated descriptions with medical concepts, but it is more than just a simple matching of UGTs against a vocabulary.
We call the task of mapping the language of UGTs to medical terminology \textit{medical concept normalization}. 
It is especially difficult since in social media, patients discuss different concepts of illness and a wide array of drug reactions. Moreover, UGTs from social networks are typically ambiguous and very noisy, containing misspelled words, incorrect grammar, hashtags, abbreviations, smileys, different variations of the same word, and so on. 


Traditional approaches for concept normalization utilized lexicons and knowledge bases with string matching. The most popular knowledge-based system for mapping texts to UMLS identifiers is MetaMap \cite{aronson2001effective}. 
This linguistic-based system uses lexical lookup and variants by associating a score with phrases in a sentence. The state-of-the-art baseline for clinical and scientific texts is DNorm \cite{leaman2013dnorm}. 
DNorm adopts a pairwise learning-to-rank technique using vectors of query mentions and candidate concept terms. 
This model outperforms MetaMap significantly, increasing the macro-averaged F-measure by 25\% on an NCBI disease dataset. However, while these tools have proven to be effective for patient records and research papers, they achieve moderate results on social media texts~\cite{nikfarjam2015pharmacovigilance,limsopatham2016normalising}. 

Recent works go beyond string matching: these works have tried to view the problem of matching a one- or multi-word expression against a knowledge base as a supervised sequence labeling problem. \citet{limsopatham2016normalising} utilized convolutional neural networks (CNNs) for phrase normalization in user reviews, while \citet{tutubalina2018medical}, \citet{Han201749}, and \citet{Belousov201754} applied recurrent neural networks (RNNs) to UGTs, achieving similar results.
These works were among the first applications of deep learning techniques to medical concept normalization. 


The goal of this work is to study the use of deep neural models, i.e., contextualized word representation model BERT \cite{devlin2018bert} and Gated Recurrent Units (GRU)~\citep{cho-EtAl:2014:EMNLP2014} with an attention mechanism, paired with \textit{word2vec} word embeddings and contextualized ELMo embeddings \cite{Peters:2018}. We investigate if a joint architecture with special provisions for domain knowledge can further improve the mapping of entity mentions from UGTs to medical concepts. We combine the representation of an entity mention constructed by a neural model and distance-like similarity features using vectors of an entity mention and concepts from the UMLS. 
We experimentally demonstrate the effectiveness of the neural models for medical concept normalization on three real-life datasets of tweets and user reviews about medications with two evaluation procedures.





\section{Problem Statement}\label{sec:problem}

\begin{table}[t!]\centering
\setlength{\tabcolsep}{3pt}
\begin{tabular}{|p{4cm}|p{3cm}|}
\hline \textbf{Entity from UGTs} & \textbf{Medical Concept} \\ \hline
no sexual interest&Lack of libido  \\\hline
nonsexual being&Lack of libido\\\hline
couldnt remember long periods of time or things&Poor long-term memory\\\hline
loss of memory& Amnesia\\\hline
bit of lower back pain & Low Back Pain \\\hline
pains & Pain\\\hline
like i went downhill&Depressed mood\\\hline
just lived day by day&Apathy\\ \hline
dry mouth&Xerostomia\\
\hline
\end{tabular}

\caption{\label{tab:examples} Examples of extracted social media entities and their associated medical concepts. }
\end{table}

Our main research problem is to investigate the content of UGTs with the aim to learn the transition between a layperson’s language and formal medical language. Examples from Table~\ref{tab:examples} show that an automated model has to account for the semantics of an entity mention. For example, it has to be able to map not only phases with shared $n$-grams \textit{no sexual interest} and \textit{nonsexual being} into the concept ``Lack of libido'' but also separate the phase \textit{bit of lower back pain} from the broader concept ``Pain'' and map it to a narrower concept.

While focusing on user-generated texts on social media, in this work we seek to answer the following research questions.

\begin{enumerate}[start=1,label={\bfseries RQ\arabic*:},leftmargin=3\parindent]
\item Do distributed representations reveal important features for medication use in user-generated texts?
\item  Can we exploit the semantic similarity between entity mentions from user comments and medical concepts? Do the neural models produce better results than the existing effective baselines? [current research]
\item  How to integrate linguistic knowledge about concepts into the models? [current research]
\item How to jointly learn concept embeddings from UMLS and representations of health-related entities from UGTs? [future research]
\item How to effectively use of contextual information to map entity mentions to medical concepts? [future research]
\end{enumerate}

To answer RQ1, we began by collecting UGTs from popular medical web portals and investigating distributed word representations trained on 2.6 millions of health-related user comments. In particular, we analyze drug name representations using clustering and chemoinformatics approaches. The analysis demonstrated that similar word vectors correspond to either drugs with the same active compound or to drugs with close therapeutic effects that belong to the same therapeutic group. It is worth noting that chemical similarity in such drug pairs was found to be low. Hence, these representations can help in the search for compounds with potentially similar biological effects among
drugs of different therapeutic groups \cite{tutubalina2017using}. 

To answer RQ2 and RQ3, we develop several models and conduct a set of experiments on three benchmark datasets where social media texts are extracted from user reviews and Twitter. We present this work in Sections~\ref{sec:models} and \ref{sec:exp}. We discuss RQ4 and RQ5 with research plans in Section~\ref{sec:future}.

\section{Methods} \label{sec:models}
Following state-of-the-art research \cite{limsopatham2016normalising,sarker2018data}, we view concept normalization as a classification problem. 

To answer RQ2, we investigate the use of neural networks to learn the semantic representation of an entity before mapping its representation to a medical concept. First, we convert each mention into a vector representation using one of the following (well-known) neural models:

\begin{enumerate}[(1)]
\item bidirectional LSTM \cite{lstm97and95} or GRU \cite{cho-EtAl:2014:EMNLP2014} with an attention mechanism and a hyperbolic tangent activation function on top of 200-dimensional word embeddings obtained to answer RQ1;
\item a bidirectional layer with attention on top of deep contextualized word representations ELMo \cite{Peters:2018};
\item a contextualized word representation model BERT~\cite{devlin2018bert}, which is a multi-layer bidirectional Transformer encoder.
\end{enumerate} 
We omit technical explanations of the neural network architectures due to space constraints and refer to the studies above.

Next, the learned representation is concatenated with a number of semantic similarity features based on prior knowledge from the UMLS Metathesaurus. Lastly, we add a softmax layer to convert values to conditional probabilities. 

The most attractive feature of the biomedical domain is that domain knowledge is prevailing in this domain for dozens of languages. In particular, UMLS is undoubtedly the largest lexico-semantic resource for medicine, containing more than 150 lexicons with terms from 25 languages. To answer RQ3, we extract a set of features to enhance the representation of phrases. These features contain cosine similarities between the vectors of an input phrase and a concept in a medical terminology dictionary. We use the following strategy, which we call~\textsc{TF-IDF (max)}, to construct representations of a concept and a mention: represent a medical code as a set of terms; for each term, compute the cosine distance between its TF-IDF representation and the entity mention; then choose the term with the largest similarity.

\section{Experiments}\label{sec:exp}
We perform an extensive evaluation of neural models on three datasets of UGTs, namely \textbf{CADEC}~\cite{karimi2015cadec}, \textbf{PsyTAR}~\cite{zolnoori2019systematic}, and \textbf{SMM4H 2017} \cite{sarker2018data}. The basic task is to map a social media phrase to a relevant medical concept. 

\subsection{Data}



\paragraph*{CADEC.} 
CSIRO Adverse Drug Event Corpus (CADEC)~\cite{karimi2015cadec} is the first richly annotated and
publicly available corpus of medical forum posts taken from \emph{AskaPatient}\footnote{\url{https://www.askapatient.com}}. 
This dataset contains 1253 UGTs about 12 drugs divided into two categories: Diclofenac and Lipitor. 
All posts were annotated manually for 5 types of entities: ADR, Drug, Disease, Symptom, and Finding. 
The annotators performed terminology association using the Systematized
Nomenclature Of Medicine Clinical Terms (SNOMED CT). 
We removed ``conceptless'' or ambiguous mentions for the purposes of evaluation. There were 6,754 entities and 1,029 unique codes in total. 

\paragraph*{PsyTAR.} 
Psychiatric Treatment Adverse Reactions (PsyTAR) corpus \cite{zolnoori2019systematic} is the second open-source corpus of user-generated posts taken from AskaPatient. This dataset includes 887 posts about four psychiatric medications from two classes: (i) Zoloft and Lexapro from the Selective Serotonin Reuptake Inhibitor (SSRI) class and (ii) Effexor and Cymbalta from the Serotonin Norepinephrine Reuptake Inhibitor (SNRI) class. All posts were annotated manually for 4 types of entities: ADR, withdrawal symptoms, drug indications, and sign/symptoms/illness. 
The corpus consists of 6556 phrases mapped to 618 SNOMED codes.


\paragraph*{SMM4H 2017.} 
In 2017, \citet{sarker2018data} organized the Social Media Mining for Health (SMM4H) shared task which introduced a dataset with annotated ADR expressions from \emph{Twitter}. Tweets were collected using 250 keywords such as generic and trade names for medications along with misspellings. Manually extracted ADR expressions were mapped to Preferred Terms (PTs) of the Medical Dictionary for Regulatory Activities (MedDRA). The training set consists of 6650 phrases mapped to 472 PTs. The test set consists of 2500 mentions mapped to 254 PTs. 


\subsection{Evaluation Details}
We evaluate our models based on classification accuracy, averaged across randomly divided five folds of the CADEC and PsyTAR corpora. For SMM4H 2017 data, we adopted the official training and test sets \cite{sarker2018data}. Analysis of randomly split folds shows that \textit{Random KFolds} create a high overlap of expressions in exact matching between subsets (see the baseline results in Table~\ref{tab:res}). 
Therefore, we set up a specific train/test split procedure for 5-fold cross-validation on the CADEC and PsyTAR corpora: we removed duplicates of mentions and grouped medical records we are working with into sets related to specific medical codes. Then, each set has been split independently into $k$ folds, and all folds have been merged into the final $k$ folds named \textit{Custom KFolds}. Random folds of CADEC are adopted from \cite{limsopatham2016normalising} for a fair comparison. Custom folds of CADEC are adopted from our previous work \cite{tutubalina2018medical}. PsyTAR folds are available on Zenodo.org\footnote{\url{https://doi.org/10.5281/zenodo.3236318}}.
%
We have also implemented a simple \textit{baseline} approach that uses exact lexical matching with lowercased annotations from the training set.

\subsection{Results}

Table \ref{tab:res} shows our results for the concept normalization task on the Random and Custom KFolds of the CADEC, PsyTAR, and SMM4H 2017 corpora.

To answer RQ2, we compare the performance of examined neural models with the baseline and state-of-the-art methods in terms of accuracy. Attention-based GRU with ELMo embeddings showed improvement over GRU with \textit{word2vec} embeddings, increasing the average accuracy to 77.85 (+3.65). The semantic information of an entity mention learned by BERT helps to improve the mapping abilities, outperforming other models (avg. accuracy 83.67). Our experiments with recurrent units showed that GRU consistently outperformed LSTM on all subsets, and attention mechanism provided further quality improvements for GRU. From the difference in accuracy on the Random and Custom KFolds, we conclude that future research should focus on developing extrinsic test sets for medical concept normalization. In particular, the BERT model's accuracy on the CADEC Custom KFolds decreased by 9.23\% compared to the CADEC Random KFolds. 

To answer RQ3, we compare the performance of models with additional similarity features (marked by ``w/'') with others. Indeed, joint models based on GRU and similarity features gain 2-5\% improvement on sets with Custom KFolds. The joint model based on BERT and similarity features stays roughly on par with BERT on all sets. We also tested different strategies for constructing representations using word embeddings and TF-IDF for all synonyms' tokens that led to similar improvements for GRU.

\begin{table*}[]
\centering
\begin{tabular}{|l|c|c|c|c|c|}
\hline
\multirow{2}{*}{\textbf{Method}} & \multicolumn{2}{c|}{\textbf{CADEC}} & \multicolumn{2}{c|}{\textbf{PsyTAR}} & \multicolumn{1}{c|}{\textbf{SMM4H}} \\ \cline{2-6} 
 & \multicolumn{1}{c|}{Random} & \multicolumn{1}{c|}{Custom} & \multicolumn{1}{c|}{Random} & \multicolumn{1}{c|}{Custom} & \multicolumn{1}{c|}{Official} \\ \hline
Baseline: match with training set annotation &  66.09 & 0.0 & 56.04 & 2.63 & 67.12 \\ \hline
DNorm \cite{limsopatham2016normalising} & 73.39 & - & - & -  & - \\
CNN \cite{limsopatham2016normalising} & 81.41 & - & - & -  & - \\ 
RNN \cite{limsopatham2016normalising} & 79.98  & - & - & -  & - \\ 

Attentional Char-CNN  \cite{niu2018multi} & 84.65 & - & - & - & - \\
Hierarchical Char-CNN \cite{Han201749} & - & - & - & - & 87.7 \\

Ensemble \cite{sarker2018data}& - &-  &-  & - & 88.7 \\
\hline
GRU+Attention  & 82.19 & 66.56  & 73.12 &  65.98 &  83.16\\ 
GRU+Attention w/ \textsc{TF-IDF (max)}& 84.23 & 70.05  & 75.53 & 68.59 &  86.28 \\  \hline
ELMo+GRU+Attention & 85.06 &  71.68  & 77.58 & 68.34 & 86.60 \\ 
ELMo+GRU+Attention w/ \textsc{TF-IDF (max)} & 85.71 & 74.70  & 79.52 & 70.05 & 87.52 \\  \hline
BERT &  88.69 & 79.83 & 83.07 & 77.52 & 89.28 \\ 
BERT w/ \textsc{TF-IDF (max)} & 88.84 & 79.25 & 82.37 & 77.33 &  89.64 \\ \hline
\end{tabular}
\caption{\label{tab:res} The performance of the proposed models and the state-of-the-art methods in terms of accuracy.}
\end{table*}
\section{Future Directions}\label{sec:future}

\paragraph*{RQ4.} 
Future research might focus on developing an embedding method that jointly maps extracted entity mentions and UMLS concepts into the same continuous vector space. The methods could help us to easily measure the similarity between words and concepts in the same space. Recently, \citet{yamada2016joint} demonstrated that co-trained vectors improve the quality of both word and entity representations in entity linking (EL) which is a task closely related to concept normalization. We note that most of the recent EL methods focus on the disambiguation sub-task, applying simple heuristics for candidate generation. The latter is especially challenging in medical concept normalization due to a significant language difference between medical terminology and patient vocabulary.



\paragraph*{RQ5.} 
Error analysis has confirmed that models often misclassify closely related concepts (e.g., ``Emotionally detached''  and ``Apathy'') and antonymous concepts (e.g., ``Hypertension'' and ``Hypotension''). We suggest to take into account not only the distance-like similarity between entity mentions and concepts but the mention's context, which is not used directly in recent studies on concept normalization. The context can be represented by the set of adjacent words or entities. As an alternative, one can use a conditional random field (CRF) to output the most likely sequence of medical concepts discussed in a review.

\section{Related Work}

In 2004, the research community started to address the needs to automatically detect biomedical entities in free texts through shared tasks. \citet{huang2015community} survey the work done in the organization of biomedical NLP (BioNLP) challenge evaluations up to 2014. These tasks are devoted to the normalization of \begin{inparaenum}[(1)]
\item genes from scientific articles (BioCreative I-III in 2005-2011);
\item chemical entity mentions (BioCreative IV CHEMDNER in 2014);
\item disorders from abstracts (BioCreative V CDR Task in 2015);
\item diseases from clinical reports (ShARe/CLEF eHealth 2013; SemEval 2014 task 7).
\end{inparaenum} 
Similarly, the \emph{CLEF Health} 2016 and 2017 labs addressed the problem of ICD coding of free-form death certificates (without specified entity mentions). Traditionally, linguistic approaches based on dictionaries, association measures, and syntactic properties have been used to map texts to a concept from a controlled vocabulary~\cite{aronson2001effective,van2016erasmus,mottin2016bitem,ghiasvand2014uwm,tang2014uth}. \citet{leaman2013dnorm} proposed the DNorm system based on a pairwise learning-to-rank technique using vectors of query mentions and candidate concept terms. These vectors are obtained from a tf-idf representation of all tokens from training mentions and concept terms. \citet{zweigenbaum2016hybrid} utilized a hybrid method combining simple dictionary projection and mono-label supervised classification from ICD coding. Nevertheless, the majority of biomedical research on medical concept extraction primarily focused on scientific literature and clinical records~\cite{huang2015community}. \citet{zolnoori2019systematic} applied a popular dictionary look-up system cTAKES on user reviews. 
cTAKES based on additional PsyTAR's dictionaries achieves twice better results (0.49 F1 score on the exact matching). 
Thus, dictionaries gathered from layperson language can efficiently improve automatic performance.

The 2017 SMM4H shared task \cite{sarker2018data} was the first effort for the evaluation of NLP methods for the normalization of health-related text from social media on publicly released data. Recent advances in neural networks have been utilized for concept normalization: recent studies have employed convolutional neural networks \cite{limsopatham2016normalising,niu2018multi} and recurrent neural networks \cite{Belousov201754,Han201749}. These works have trained neural networks from scratch using only entity mentions from training data and pre-trained word embeddings. 
To sum up, most methods have dealt with encoding information an entity mention itself, ignoring the broader context where it occurred. Moreover, these studies did not examine an evaluation methodology tailored to the task.

\section{Conclusion}\label{sec:concl}

In this work, we have performed a fine-grained evaluation of neural models for medical concept normalization tasks. We employed several powerful models such as BERT and RNNs paired with pre-trained word embeddings and ELMo embeddings. We also developed a joint model that combines (i) semantic similarity features based on prior knowledge from UMLS and (ii) a learned representation that captures extensional semantic information of an entity mention. We have carried out experiments on three datasets using 5-fold cross-validation in two setups. Each dataset contains phrases and their corresponding SNOMED or MedDRA concepts. Analyzing the results, we have found that similarity features help to improve mapping abilities of joint models based on recurrent neural networks paired with pre-trained word embeddings or ELMo embeddings while staying roughly on par with the advanced language representation model BERT in terms of accuracy. Different setups of evaluation procedures affect the performance of models significantly: the accuracy of BERT is 7.25\% higher on test sets with a simple random split than on test sets with the proposed custom split. Moreover, we have discussed some interesting future research directions and challenges to be overcome.

\section*{Acknowledgments}
We thank Sergey Nikolenko for helpful discussions. This research was supported by the Russian Science Foundation grant no. 18-11-00284.

\bibliography{acl2019}
\bibliographystyle{acl_natbib}

\end{document}